\documentclass[letterpaper, 10 pt, conference]{ieeeconf}
\IEEEoverridecommandlockouts
\usepackage[font=small,labelfont=bf]{caption}

\usepackage[utf8]{inputenc}
\usepackage[T1]{fontenc}
\usepackage{graphicx}
\usepackage{amsmath,amssymb,bm}
\usepackage{booktabs}
\usepackage{multirow}
\usepackage{xspace}
\usepackage[export]{adjustbox}
\usepackage{subcaption}
\usepackage{cite}
\usepackage{stfloats}

\usepackage[colorlinks=true,linkcolor=black,citecolor=black,urlcolor=RoyalBlue]{hyperref}
\usepackage[dvipsnames]{xcolor} % for the named color 'RoyalBlue'
\usepackage{url}

\newcommand{\projname}{Genie Centurion\xspace}
\newcommand{\shortname}{GCENT\xspace}

\title{\LARGE \bf
Genie Centurion: Accelerating Scalable Real-World Robot Training with Human Rewind-and-Refine Guidance
}

\author{%
Wenhao Wang$^{*}$ \quad Jianheng Song$^{*}$ \quad Chiming Liu$^{*}$ \quad Jiayao Ma \\
Siyuan Feng \quad Jingyuan Wang \quad Yuxin Jiang \quad Kylin Chen \\
Sikang Zhan \quad Yi Wang \quad Tong Meng \quad Modi Shi \quad Xindong He \\
Guanghui Ren \quad Yang Yang \quad Maoqing Yao \\
\small \textbf{AgiBot} \\
\small \href{https://genie-centurion.github.io}{\color{RoyalBlue}\textbf{genie-centurion.github.io}} \\
\thanks{$^{*}$ Equal contribution.}%
\thanks{All authors are with AgiBot.}%
}

\begin{document}

\twocolumn[{%
  \renewcommand\twocolumn[1][]{#1}%
  \maketitle
  \vspace{-3mm}
  \begin{center}
    \includegraphics[width=\textwidth]{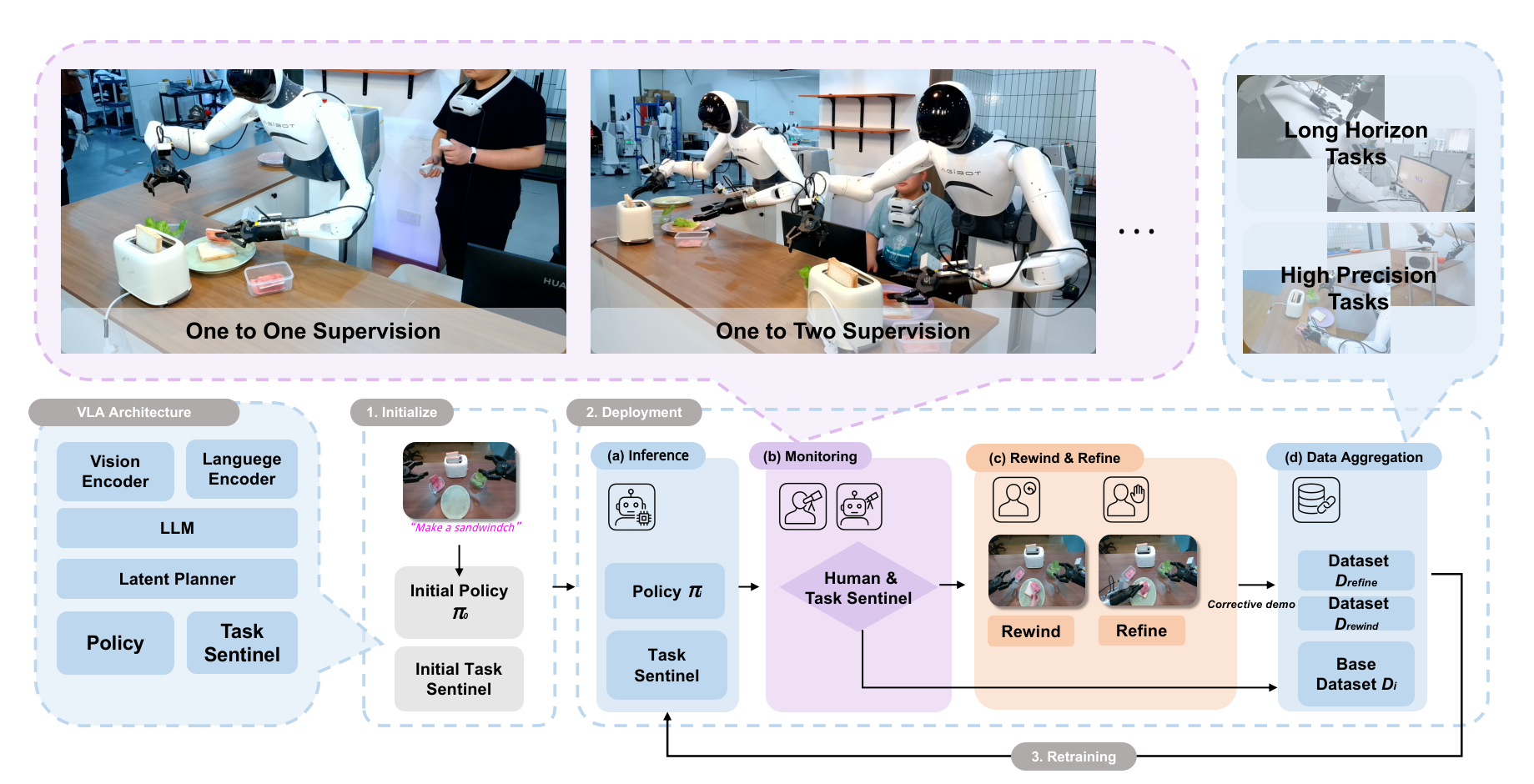}
    \captionsetup{type=figure}
    \caption{\textbf{GCENT System} is a human-in-the-loop scalable data collection system. By leveraging the VLA policy inference and the rewind \& refine teleoperation, it allows the policy to improve itself in deployment. Therefore, imperfect policies can be deployed, with one human supervising multiple robots.}
    \label{fig:framework}
  \end{center}
  \vspace{-2mm}
}]

%===============================================================================

\begin{abstract}
    While Vision-Language-Action (VLA) models show strong generalizability in various tasks, real-world deployment of robotic policy still requires large-scale, high-quality human expert demonstrations. However, data collection via human teleoperation requires continuous operator attention, which is costly, hard to scale. To address this, we propose \projname (\shortname), a scalable and general data collection paradigm based on human rewind-and-refine guidance, enabling robots' interactive learning in deployment. \shortname starts at an imperfect policy and improves over time. When the robot execution failures occur, \shortname allows robots to revert to a previous state with a rewind mechanism, after which a teleoperator provides corrective demonstrations to refine the policy. This framework supports a one-human-to-many-robots supervision scheme with a Task Sentinel module, which autonomously predicts task success and solicits human intervention when necessary. Empirical results show that \shortname achieves up to 40\% higher task success rates than state-of-the-art data collection methods, and reaches comparable performance using less than half the data in long-horizon and precise tasks. We also quantify the data yield-to-effort ratio under multi-robot scenarios, demonstrating \shortname's potential for scalable and cost-efficient robot policy training in real-world environments.
    
\end{abstract}

%===============================================================================

\section{Introduction}

In robotics, large models have recently shown great potential, especially Vision-Language-Action (VLA) models \cite{rt22023arxiv, octo_2023, kim24openvla, liu2024rdt, wen2024tinyvla, qu2025spatialvlaexploringspatialrepresentations}. Training such models requires large amounts of human demonstration data for imitation learning. The predominant approach is full human teleoperation \cite{wang2024dexcap, chi2024universal, wu2024fastumi}, in which operators must continuously teleoperate robots to provide demonstrations. This direct coupling between human attention and robot operation time inherently constrains collection efficiency to roughly a 1:1 ratio. Efficiency is usually much lower than 1:1, considering additional overhead from environment setup, task resets, and human errors, significantly reducing the net usable data collected. This raises a natural question: can we reduce operator cognitive load to scale one-to-many data collection to make the efficiency larger than 1:1?

Previous research investigates scaling the data synthesis in simulation \cite{jiang2024dexmimicen, mandlekar2023mimicgen, mu2025robotwin}. However, sim-to-real methods often suffer from the simulation gap, especially in contact-rich and high-precision manipulation tasks, where reliable transfer is not guaranteed. In contrast, scaling the data collection in the real-world robots is economically inefficient, especially the “collect-then-deploy” paradigm, because the full cost of data acquisition occurs before real-world utility is realized. 

Autonomous driving research \cite{DBLP:journals/corr/abs-1810-02890} has demonstrated the effectiveness of a “deploy-then-collect” paradigm, where imperfect models are deployed early, data from their real-world operation is streamed back, and policies are iteratively improved. The Dataset Aggregation (DAgger) algorithm \cite{DBLP:journals/corr/abs-1011-0686} formalizes a similar principle in learning: query expert corrections in the states actually visited by the current policy, thereby focusing collection on informative regions of the state space. With a human in the loop to safeguard task success, this approach enables a continuous closed loop of deployment and improvement.

In robotic manipulation tasks, we observe, policy failures often tend to cluster around specific, high-impact decision points, for example, the last action chunk of grasping, pressing, inserting, etc. While reaching the object is often easy. Conventional teleoperation collects full-trajectory data, captures large volumes of redundant or low-value data. A more targeted strategy—intervening only at these critical states—could concentrate data collection where it most effectively improves policy performance.

Given the challenges and observations, we propose \shortname, a DAgger-inspired data collection framework tailored for efficient real-world robotic policy learning. In this paradigm, the human operator acts as a guardian, intervening only when the policy fails or is about to fail. \shortname introduces a rewind mechanism that allows the operator to revert the robot to a previous state, thereby enhancing the diversity and coverage of critical state space. Additionally, \shortname incorporates a Task Sentinel module, a vision-language-based model designed to autonomously predict task success and request human intervention when necessary. This reduces reliance on continuous human attention, therefore, enabling scalable data collection. 

Our primary contributions include:

1. We introduce \projname (\shortname), a unified framework designed for scalable robot policy training through interventions triggered by failures, complemented by a rewind mechanism to enhance state-space coverage.

2. We perform extensive real-world experiments comparing \shortname to conventional teleoperation-based data collection methods, demonstrating significant improvements in task success rates and substantially reduced human operational effort.

3. We propose the Task Sentinel module, demonstrating that Task Sentinel enables scalable supervision, allowing a single operator to effectively oversee multiple robots simultaneously.

%===============================================================================

\section{Related Work}
\label{sec:relates}
\subsection{Learning-Based Manipulation and the Role of Data}

Learning-based approaches have demonstrated promising capabilities for robotic manipulation, particularly in multi-task, multi-modal, and open-ended instruction-following scenarios. Most existing methods adopt a two-stage training paradigm: large-scale pre-training followed by task-specific fine-tuning\cite{kim2025finetuningvisionlanguageactionmodelsoptimizing, singh2024studyoptimizationsfinetuninglarge}. In this setting, the performance heavily depends on the quality and diversity of the fine-tune data. Recent studies indicate that merely increasing data quantity is not sufficient for robust generalization; instead, factors such as coverage of failure cases, and diversity of environments and objects have a greater impact \cite{lin2024data_scaling, lin2024data}. To address this, some researchers construct standard cross-embodiment datasets on various tasks \cite{wu2024robomind, agibot2025colosseo}, while others explore efficient collection strategies, such as compressing diverse spatial and linguistic exploration into minimal demonstrations \cite{huang2025adversarial}. However, most approaches still focus on repetitive collecting full trajectories and lack mechanisms to capture policy failed states systematically, limiting policy performance convergence efficiency.

\subsection{Interactive Imitation Learning and Human-in-the-loop Supervision}

To address the distribution shift in behavior cloning (BC) \cite{zhao2023learningfinegrainedbimanualmanipulation, chi2024diffusionpolicyvisuomotorpolicy}, DAgger  \cite{DBLP:journals/corr/abs-1011-0686} introduces expert supervision during rollouts to iteratively aggregate the data distribution. SafeDAgger and LazyDAgger \cite{zhang2016queryefficientimitationlearningendtoend, hoque2021lazydagger} reduce human burden through safety prediction and switching cost modeling. HG-DAgger \cite{DBLP:journals/corr/abs-1810-02890} uses a human expert as a gating function and maintains exclusive control authority. EnsembleDAgger \cite{DBLP:journals/corr/abs-1807-08364} leverages an ensemble-based uncertainty estimation, while ThriftyDAgger  \cite{hoque2021thriftydagger} learns a budget-aware switching policy to trigger human interventions at high-risk states. While these works improve intervention strategies, they are mostly designed for simulation or driving contexts, with limited application to real-world manipulation. HIL-SERL \cite{luo2024hilserl} achieves a 100\% success rate in short tasks by reinforcement learning. RoboCopilot \cite{wu2025robocopilothumanintheloopinteractiveimitation, xu2025hactshumanascopilotteleoperationrobot} emphasizes fluent control transfer, but still relies on constant human monitoring, limiting its scaling in the real world. Fleet-DAgger and SIRIUS-FLEET \cite{hoque2022fleetdagger, liu2024siriusfleet} extend these strategies to multi-robot systems with supervision scheduling and prediction. We extend this research on real-world complex tasks such as bimanual, long-horizon, and high-precision tasks.
In contrast to recent approaches using VQA-style success checking \cite{Luo_2024, ma2024generative,du2023visionlanguagemodelssuccessdetectors}, we adopt a reward model-style value predictor head as Task Sentinel that independently determines whether each step has succeeded.

%===============================================================================

\section{GCENT Data Collection System}

\begin{figure}[t]
    \centering
    \includegraphics[width=3in]{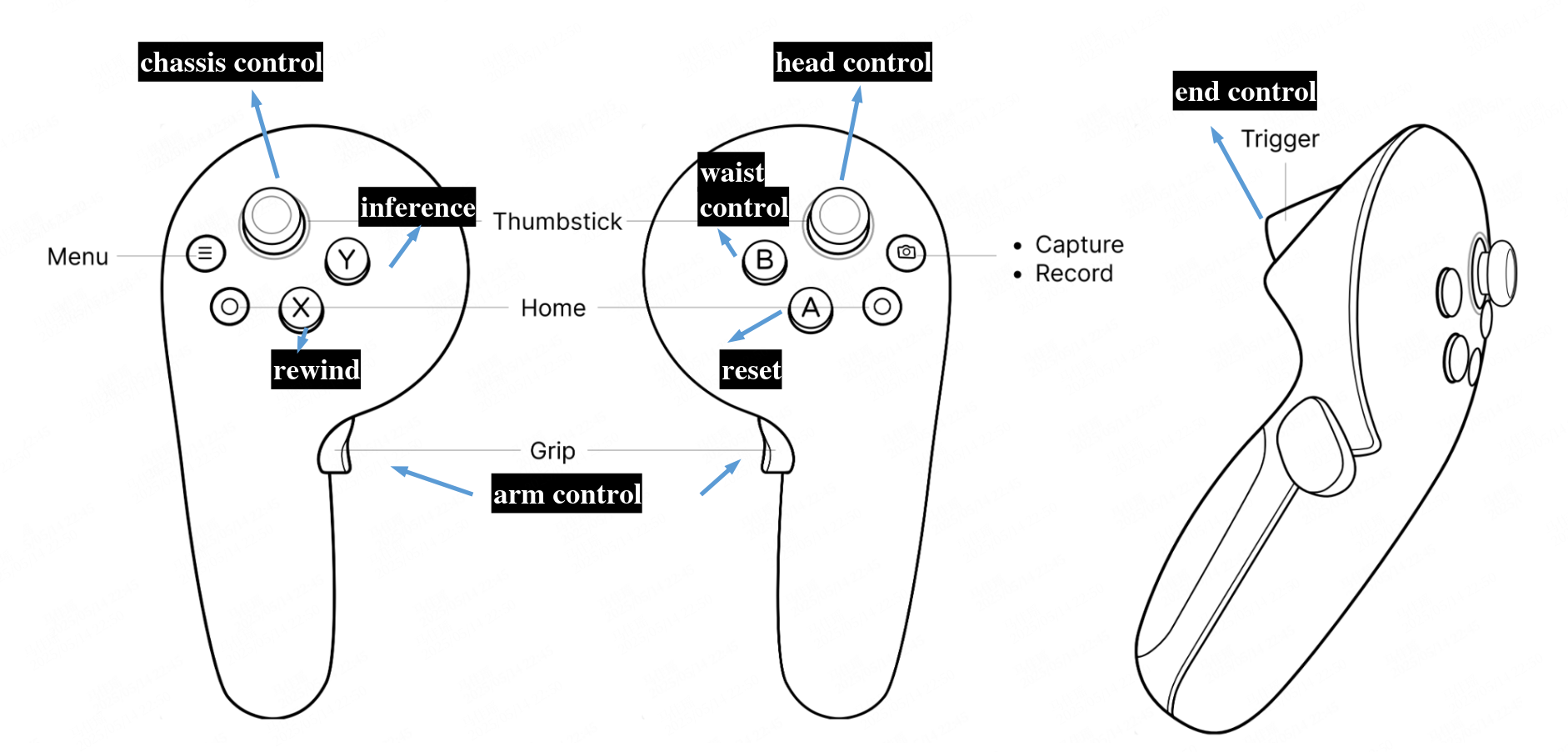}
    \caption{\textbf{Teleoperation Setup} illustrating button assignments for the dual 6-DoF controllers and their functions in GCENT. }
    \vspace{-10pt}
    \label{fig:vr}
\end{figure}

\begin{figure*}[!t]
    \centering
    \includegraphics[width=\textwidth]{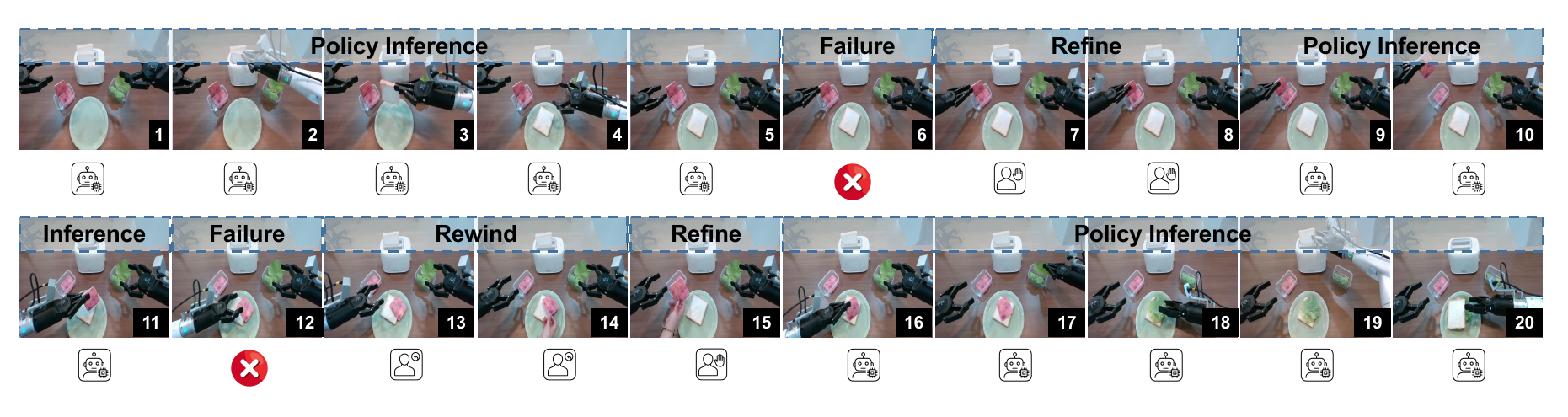}
    \caption{\textbf{Labels in sandwich assembly task}. 
    It shows the mode transitions among four core interaction modes: ``Refine'', ``Rewind'', ``Inference'' and ``Awaiting Intervention''(failure).}
    \vspace{-10pt}
    \label{fig:rewindrefine}
\end{figure*}

\label{sec:method}
This section introduces the hardware settings and the processing pipeline. Core functionalities like Task Sentinel, interactive rewind, and the refine mechanism are discussed in detail.

\subsection{Hardware Setup}

The GCENT data collection system is built upon the AgiBot G01\cite{agibot2025colosseo} robotic platform. The operator uses VR controllers to control the robot's dual-arm end-effectors and uses buttons to change between inference, intervention, and rewind modes. 

As shown in Figure \ref{fig:vr}, each button is assigned a function. The Y button initiates the inference mode, starts policy execution on the robot. The X button triggers the rewind mode, restoring the system to a previous time point. The side button initiates the intervention mode, enabling manual control for human demonstration or correction. The A button resets the system, returns the robot to its initial pose, and ends the current data collection. Other buttons are assigned to control the rest of the components of the robot's body.

\subsection{Data Processing Pipeline}
The GCENT operates in a continuously iterating data loop. It refines the policy by online interaction progressively, and reduces the intervention rate by increasing the success rate. The process is illustrated in Figure~\ref{fig:framework}.

\begin{enumerate}
    \item \textbf{Initialization}: A small set of seed data, $D_0$, is collected through human teleoperation to train an initial policy $\pi_0$.
    \item \textbf{Deployment}: This stage includes four key steps:
    \begin{enumerate}
        \item \textbf{Inference}: When the operator presses the Y button, the robot enters inference mode and performs tasks autonomously using the current policy $\pi_i$ and the Task Sentinel model $\text{Sentinel}_i$.
        \item \textbf{Monitoring}: The system determines whether the current task step is complete based on both human supervision and signals from the Task Sentinel model (detailed in Section~\ref{section:gated}). If completed, it proceeds to the next step; otherwise, a rewind or intervention is requested.
        \item \textbf{Rewind and Refine}: When the operator presses the X button, the robot will rewind to previous states, detailed in Section~\ref{section:rewind} (highlighted in Figure~\ref{fig:framework}). Then the operator presses the side button, teleoperates the robot to refine or complete the task.
        \item \textbf{Data Aggregation}: After completion or correction of the task, the effective trajectory data, particularly the successful refining trajectories $D_{\text{refine}}$ from step (2c), are aggregated in the dataset $D_{i+1}$.
    \end{enumerate}
    \item \textbf{Retraining}: The updated dataset $D_{i+1}$ is then used to fine-tune both policy model and Task Sentinel model, yielding new versions $\pi_{i+1}$ and ${\text{Sentinel}_{i+1}}$. These updated models are then deployed on the robot, and the deployment cycle (step 2) is further repeated. This iterative process continues until the robot can complete tasks and reliably monitor the task status autonomously.
\end{enumerate}

Our data pipeline logs real-time data on the robot, including 3 cameras observations ($o_t$), joint states ($s_t$), policy actions ($a_t$), and instructions. This data is first saved locally in HDF5 format, then validated, and uploaded to the cloud. Cloud services further process the data, including frame alignment, step labeling, storage management, and more, before model training begins.

The GCENT data collecting system automatically logs mode labels. There are four modes: ``Intervention'', ``Rewind'', ``Inference'', and ``Awaiting Intervention''. The trajectory segments labeled ``Intervention'', denoted as $(o_{\tau}, a_{\tau}^{\text{human}})$, are automatically identified as high quality supervised samples for training $\pi_i$. The ``Awaiting Intervention'' mode is autonomously labeled by the Task Sentinel. 

\subsection{Task Sentinel: A Multimodal Large Language Model-based Robot Step Detection Model}\label{section:gated}

In addition to human supervision, we designed an autonomous mechanism, the Task Sentinel, for the robot to determine when intervention is necessary. This model, inspired by the reward model architecture in \cite{cai2024internlm2technicalreport}, takes the current image observation $o_t$ and task instruction $l_{\text{task}}$ as input at time $t$. As shown in Equation~\ref{eq:sentinel_output}, the model outputs a boolean value $z_t$ indicating the success status of the current step:
\begin{equation}
    z_t := \text{Sentinel}(o_t, l_{\text{task}}) \in \{0,1\}
    \label{eq:sentinel_output}
\end{equation}
When the step is completed, denoted as $z_t=1$, the robot automatically proceeds to the next step. If it is not completed within a predefined time $T_{\text{max}}$ (i.e., $z_t=0 \text{ and } \Delta t > T_{\text{max}}$), GCENT transitions to the ``Awaiting Intervention'' mode until a human operator intervenes, as described by the condition:

\begin{equation}
    \text{if } (z_t = 0 \text{ and } \Delta t > T_{\text{max}}), \text{ then request human intervention.}
    \label{eq:sentinel_intervention}
\end{equation}

We chose not to detect failure moments. 
This decision stems from the significant challenge of accurately identifying all kinds of failure, particularly with limited data. The types and frequencies of errors can vary at different stages of model training. In contrast, the definition of task (or sub-task) success remains clear and stable, rendering it more suitable for training robust models under the GCENT paradigm.

Tasks are decomposed into steps (e.g., making the sandwich comprises grasping bread, placing bread, grasping lettuce, placing lettuce, etc.), and the Task Sentinel monitors each step. Human annotators identify the start and end frames for each step. We designate the final second frames of a successfully executed step as completed, while all other frames as not completed. It converts the training of the Task Sentinel model into a binary classification task.

We combine both Task Sentinel and human supervision in GCENT. In human supervision, an operator monitors the robot's actions and intervenes immediately if any error occurs. As depicted in Figure~\ref{fig:framework}, during the early stages of GCENT iteration, data collectors primarily use human supervision on a single robot due to the low policy success rate. As iterations progress and model performance improves, the Task Sentinel mechanism enables a single operator to monitor multiple robots simultaneously, requesting intervention only when necessary. Task sentinel is a key factor for scaling the GCENT approach to a one-operator-multiple-robots system.

\subsection{Rewind and Refine Mechanism}\label{section:rewind}
When the Task Sentinel requests intervention, or if the operator finds the policy fails at the task, the operator can press the X button to trigger a rewind operation. The system maintains a real-time state buffer of the past 3 seconds on the robot itself. When rewind mode is initiated, the system restores the robot to the selected historical state $s_{t-k}$ by inversely sending the states as commands. After the state rewind, the operators are allowed to apply physical perturbations and provide a correction demonstration.

%===============================================================================

\section{Experiments}
\label{sec:experiment}

\begin{figure*}[!t]
    \centering
    \captionsetup[sub]{font=small,justification=centering} % 子图标题更紧凑

    \begin{subfigure}[t]{0.24\textwidth}
        \centering
        \includegraphics[max width=\linewidth, max height=0.18\textheight]{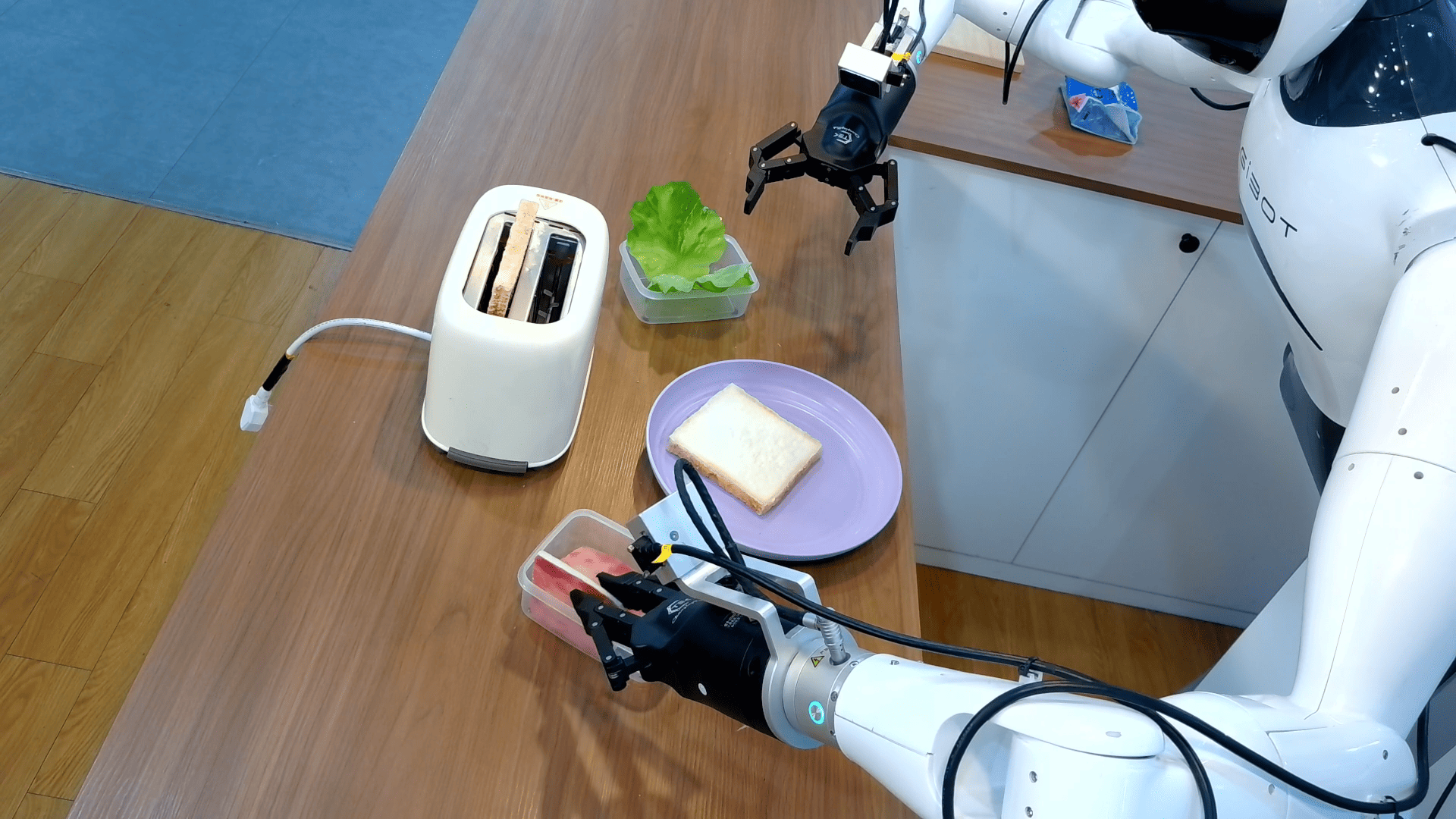}
        \caption{Sandwich Assembly}
        \label{fig:sandwich}
    \end{subfigure}\hfill
    \begin{subfigure}[t]{0.24\textwidth}
        \centering
        \includegraphics[max width=\linewidth, max height=0.18\textheight]{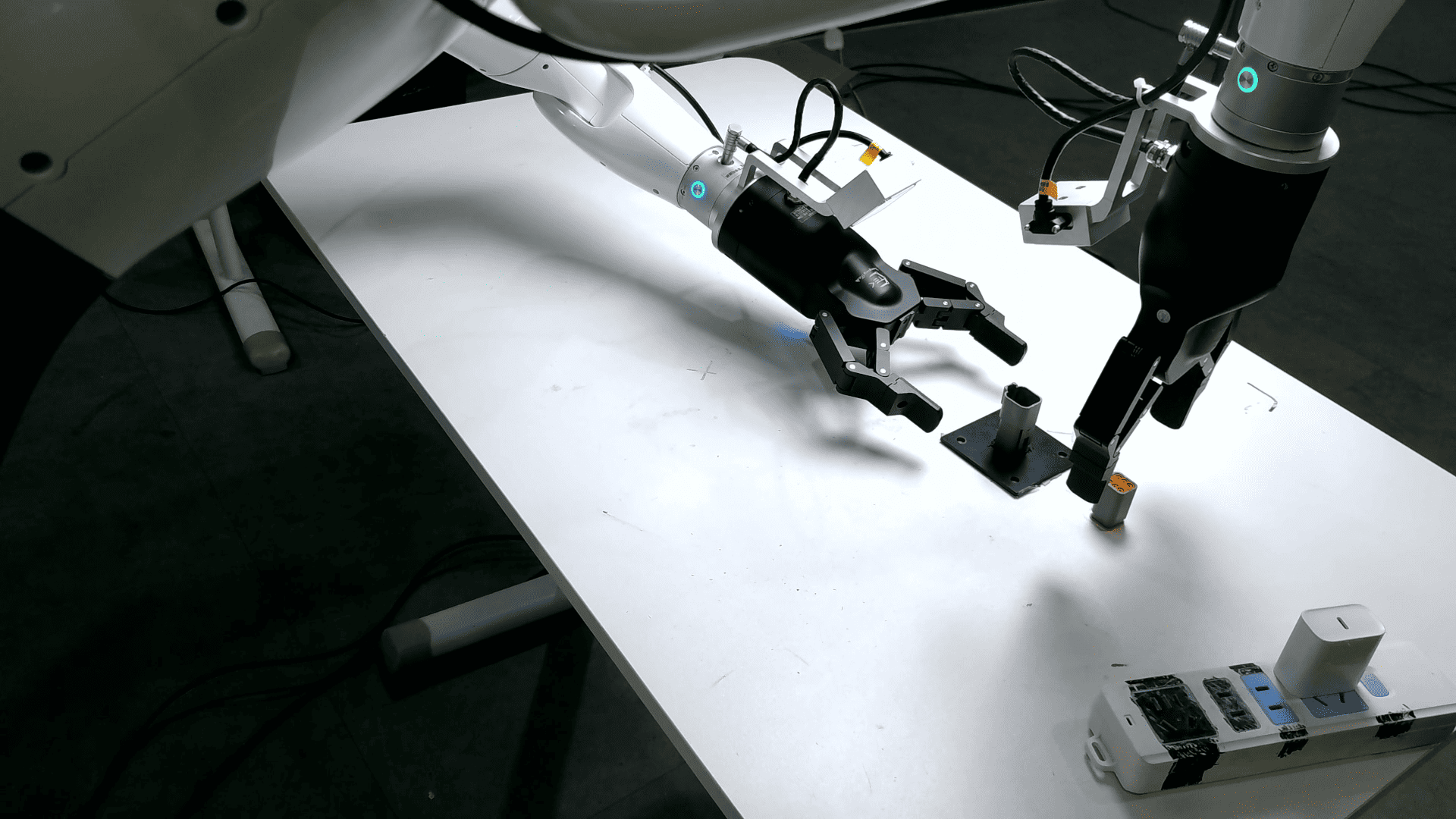}
        \caption{Connector Insertion}
        \label{fig:insert}
    \end{subfigure}\hfill
    \begin{subfigure}[t]{0.24\textwidth}
        \centering
        \includegraphics[max width=\linewidth, max height=0.18\textheight]{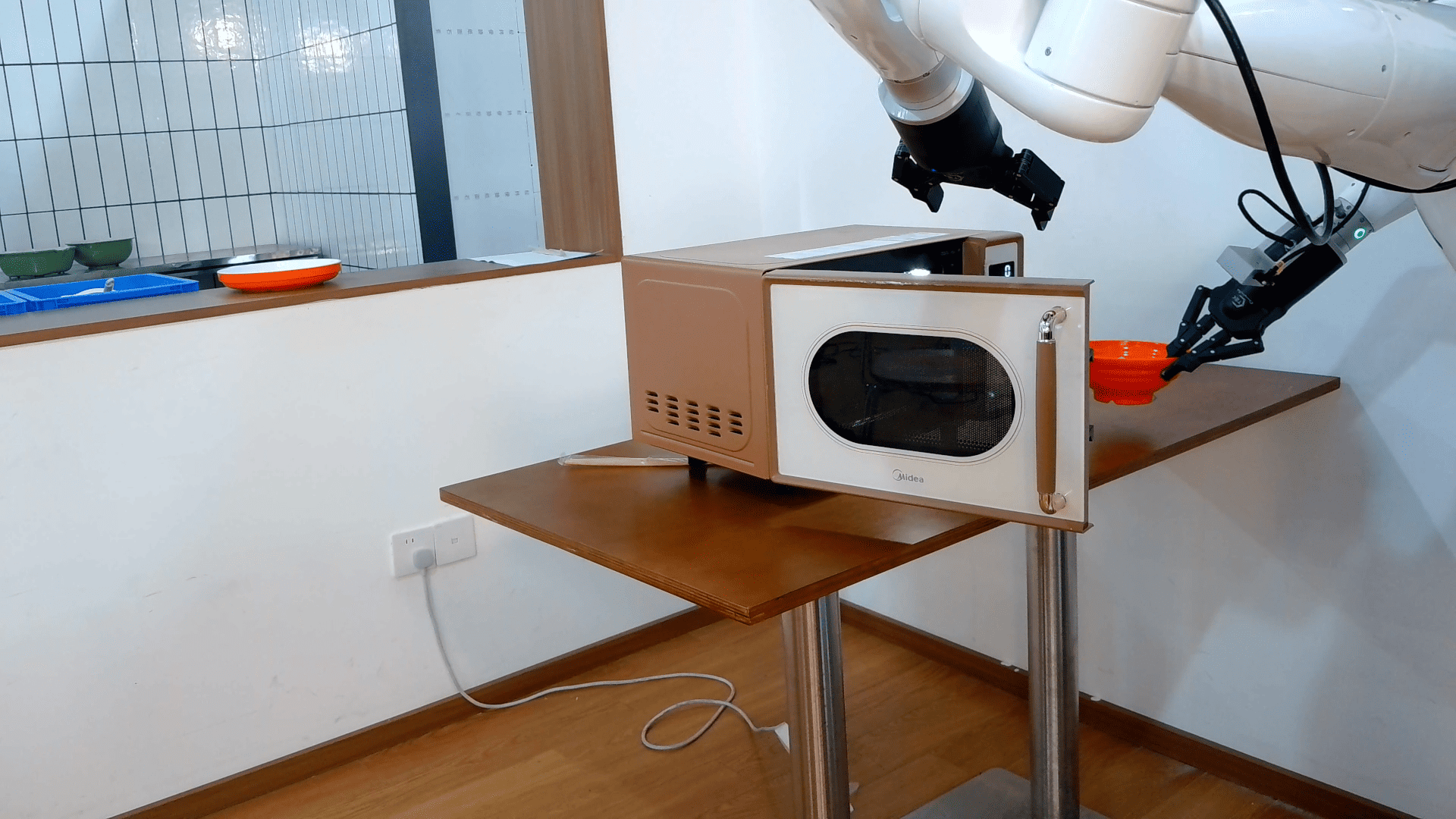}
        \caption{Microwave Heating}
        \label{fig:microwave}
    \end{subfigure}\hfill
    \begin{subfigure}[t]{0.24\textwidth}
        \centering
        \includegraphics[max width=\linewidth, max height=0.18\textheight]{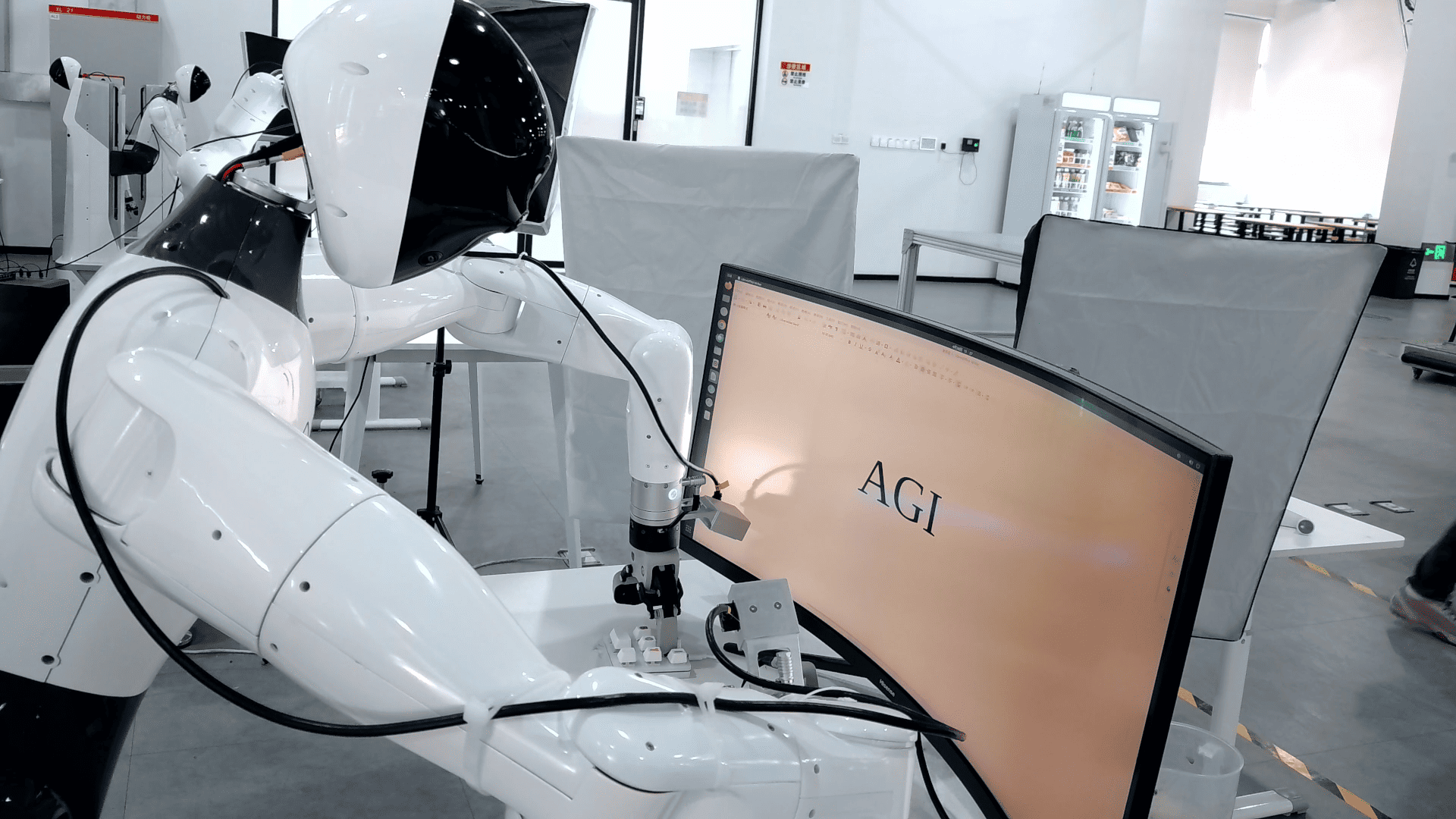}
        \caption{Typing}
        \label{fig:typing}
    \end{subfigure}

    \vspace{-4pt}
    \caption{\textbf{Four tasks in experiments.} Two long-horizon and two precise tasks.}
    \label{fig:tasks}
\end{figure*}

All experiments are in real world.

This section demonstrates that GCENT serves as an efficient and cost-effective data collection strategy, capable of achieving significant performance improvements across diverse real-world tasks with limited data. Specifically, we addresses the following questions:

\begin{itemize}
    \item \textbf{Q1:} Can GCENT achieve superior performance compared to alternative data collection strategies?
    \item \textbf{Q2:} Can GCENT improve data efficiency and reduce human supervision costs, enabling one-to-many robot supervision?
    \item \textbf{Q3:} How do different rewind strategies impact policy performance?
\end{itemize}

\subsection{Experimental Setup}

\begin{table}[ht]
\centering
\caption{Hyperparameters for policy and task sentinel training.}
\begin{tabular}{lc}
\toprule
\textbf{Hyperparameter} & \textbf{Value} \\
\midrule
Learning rate      & 2e-5 \\
Batch size         & 16 $\times$ 8 \\
Input image size   & 3$\times$224$\times$224 \\
Weight decay       & 0.01 \\
Action chunk       & 30 \\
Infer denoise timesteps  & 5 \\
Train denoise timesteps    & 1000 \\
\bottomrule
\end{tabular}
\label{tab:hyperparameters}
\end{table}

We fine-tune the main policy from the GO-1 foundation model \cite{agibot2025colosseo}. It uses the Vision-Language-Latent-Action (ViLLA) architecture, pretrained on over 1 million trajectories across 217 tasks \cite{agibot2025colosseo}. The policy infers on Agibot G1, Nvidia Orin. Given the large pre-trained data, the policy seldom fails at the approaching step and usually fails at the last action chunk. Therefore, in GCENT, the policy is finetuned using only the human intervention segments to reinforce the fragile steps.

The Task Sentinel model is built upon the InternVL 2.5 2B backbone \cite{chen2024internvl} with an additional MLP-based binary classifier head. The Task Sentinel is trained on the full dataset, as state assessment models must include both successful and failed demonstrations. 

To ensure practical relevance and task diversity, we designed four tasks based on real-world applications. Detailed training configurations are provided in the appendix.

\begin{itemize}
    \item \textbf{Sandwich Assembly}: Eight sequential bi-manual pick-and-place actions to stack ingredients into a sandwich.
    \item \textbf{Connector Insertion}: Grasp and insert a component into a tight terminal, requiring fine contact-rich control.
    \item \textbf{Microwave-Heating}: Completing a microwave heating task requires five different operations, including pull, pick, place, push, and press.
    \item \textbf{Typing}: Type out text using a small keyboard, including back-space handling for instruction-following evaluation.
\end{itemize}

 We compare three strategies under equal data volume: (1) Teleop, which is traditional human teleoperation, collecting one full trajectory at a time; (2) Adversarial Data Collection (ADC)\cite{huang2025adversarial}, which builds upon Teleop by introducing an adversarial operator who deliberately injects real-time perturbations during data collection (e.g., repositioning objects, rotating items), thereby enhancing data diversity; and (3) GCENT, our proposed interactive data collection method that adaptively addresses specific errors encountered during actual policy inference through targeted data augmentation.

For evaluation, we decompose each task into multiple steps and assign one point for each completed step, with the final performance score calculated as the ratio of completed actions to total actions (1.0 indicating complete success). This fine-grained evaluation approach provides a more nuanced performance assessment compared to binary success rates. We design 10 distinct test cases for each task and conduct 5 trials per case, reporting the average scores across all trials.

\subsection{Q1: Can GCENT Lead to Better Performance?}

\begin{table*}[t]
\caption{Comparison of average scores across data collection methodologies for four tasks.}
\centering
\renewcommand{\arraystretch}{0.9}     % 行距略微加大，提升可读性
\setlength{\tabcolsep}{4pt}            % 列间距
\footnotesize                          % IEEE 常用表格字号（也可 \scriptsize）
\resizebox{0.7\textwidth}{!}{%
\begin{tabular}{llccc}
\toprule
\textbf{Task} & \textbf{Traj.} & \textbf{Teleop} & \textbf{Adversarial} & \textbf{GCENT (Ours)} \\
\midrule
\multirow{6}{*}{Sandwich Assembly} 
& 20 & 0.28 $\pm$ 0.06 & 0.23 $\pm$ 0.04 & warm-up \\
& 40          & -               & -               & 0.69 $\pm$ 0.07 \\
& 60          & 0.30 $\pm$ 0.03 & 0.59 $\pm$ 0.14 & 0.79 $\pm$ 0.08 \\
& 80          & -               & -               & 0.76 $\pm$ 0.05 \\
& 100         & 0.45 $\pm$ 0.02 & 0.53 $\pm$ 0.15 & 0.81 $\pm$ 0.04 \\
& 120         & -               & -    & \textbf{0.91 $\pm$ 0.01} \\
\midrule
\multirow{3}{*}{Connector Insertion} 
& 20 & 0.00 $\pm$ 0.00 & 0.64 $\pm$ 0.07 & warm-up \\
& 40          & 0.10 $\pm$ 0.09 & 0.75 $\pm$ 0.08 & 0.64 $\pm$ 0.09 \\
& 60          & 0.40 $\pm$ 0.15 & 0.82 $\pm$ 0.09 & \textbf{0.90 $\pm$ 0.05} \\
\midrule
\multirow{4}{*}{Microwave-Heating} 
& 20 & 0.36 $\pm$ 0.07 & 0.48 $\pm$ 0.05 & warmup \\
& 40          & -               & -               & 0.74 $\pm$ 0.06 \\
& 60          & 0.48 $\pm$ 0.06 & 0.72 $\pm$ 0.09 & 0.89 $\pm$ 0.03 \\
& 80          & 0.55 $\pm$ 0.05 & 0.76 $\pm$ 0.11 & \textbf{0.97 $\pm$ 0.01} \\
\midrule
\multirow{5}{*}{Typing} 
& 20 & 0.11 $\pm$ 0.05 & 0.10 $\pm$ 0.07 & warm-up \\
& 40          & -               & -               & 0.26 $\pm$ 0.08 \\
& 60          & 0.18 $\pm$ 0.10 & 0.05 $\pm$ 0.03 & 0.78 $\pm$ 0.09 \\
& 80          & -               & -               & 0.85 $\pm$ 0.06 \\
& 100         & 0.10 $\pm$ 0.05 & 0.03 $\pm$ 0.02 & \textbf{0.95 $\pm$ 0.03} \\
\midrule
\multirow{3}{*}{Average} 
& 20 & 0.19 & 0.36 & warm-up \\
& 60          & 0.26 & 0.48 & \textbf{0.84} \\
& Final Round & 0.38 & 0.53 & \textbf{0.93} \\
\bottomrule
\end{tabular}
}
\label{tab:gcent_comparison}
\vspace{-4pt}
\end{table*}

We adopt a batched DAgger-style iteration \cite{wu2025robocopilothumanintheloopinteractiveimitation}. Each task starts with 20 trajectories via traditional teleoperation for warm-up, followed by 4 GCENT iterations, each of which includes 20 demonstrations, until the average score surpasses 0.9. For Sandwich, Microwave, Typing, we train policy every 40 trajectories by teleoperation and adversarial data collection, while for short task Insertion, we train every 20 trajectories. Due to the high efficiency of GCENT, we train policy every 20 trajectories for all tasks. 

We find that the warm-up model does not need to achieve strong initial performance; for instance, the Connector Insertion task starts with a score of 0, yet can still be improved to 0.9 through the GCENT method.

Table~\ref{tab:gcent_comparison} shows that GCENT achieves the highest performance across all tasks under identical data budgets. All tasks achieve an average score exceeding 0.9 within 3-5 GCENT rounds. 

Traditional teleoperation assumes all samples are equally important. However, in robotic manipulation, task success often depends on critical moments, like precise alignment or following instructions, that make up only a small part of the data but have a huge impact. These crucial samples are poorly learned by conventional methods.

Adversarial Data Collection (ADC) \cite{huang2025adversarial} addresses this limitation by introducing adversarial perturbation during the collection process, thereby improving the data distribution. However, manually crafted failures provide limited coverage and capture error patterns encountered during policy deployment. Furthermore, as policies evolve, the distribution of critical failure states shifts dynamically, reducing the effectiveness of static adversarial scenarios over time. In contrast, GCENT employs policy execution to identify critical samples, adjusting their representation in the dataset through rewind and intervention mechanisms. This approach yields a higher-quality dataset that prioritizes learning from the most informative failure states, resulting in substantial performance improvements across diverse tasks.

Empirical results show that GCENT achieves the most significant performance improvements during the initial iteration, while subsequent rounds tend to plateau or even exhibit slight regressions. This trend may be attributed to limitations in the evaluation protocol, which struggles to capture subtle improvements across iterations. This behavior reveals an important difference between artificial failure cases and real errors that arise during model execution: naturally occurring errors provide more valuable learning signals. With continued training, GCENT models are able to overcome these intermediate plateaus and ultimately converge to consistently high performance levels.

\begin{figure*}[!t]
    \centering
    \includegraphics[width=\textwidth]{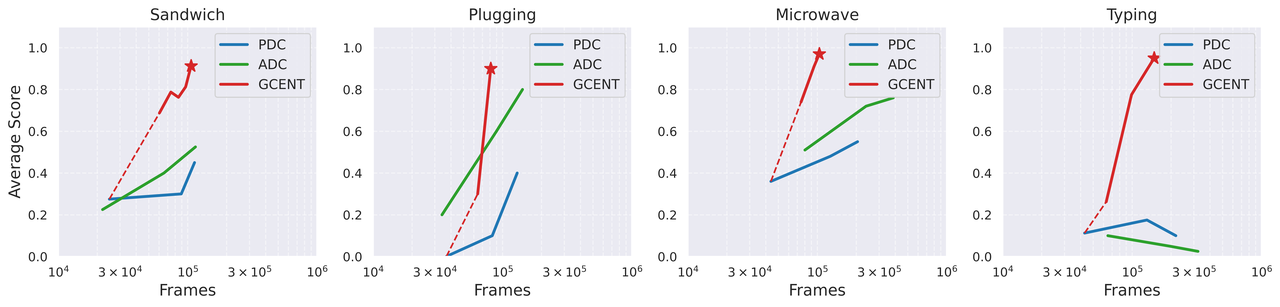}
    \caption{\textbf{Comparison of data efficiency across methods.} GCENT achieves 0.9+ task score with significantly fewer frames. At the same frame count, GCENT improves model performance by an average of 40\%; at the same performance level, GCENT requires only 44.5\% of the frames compared to teleop on average.}
    \label{fig:efficiency}
\end{figure*}

\begin{figure}[!t]
    \centering
    \includegraphics[width=\linewidth, keepaspectratio]{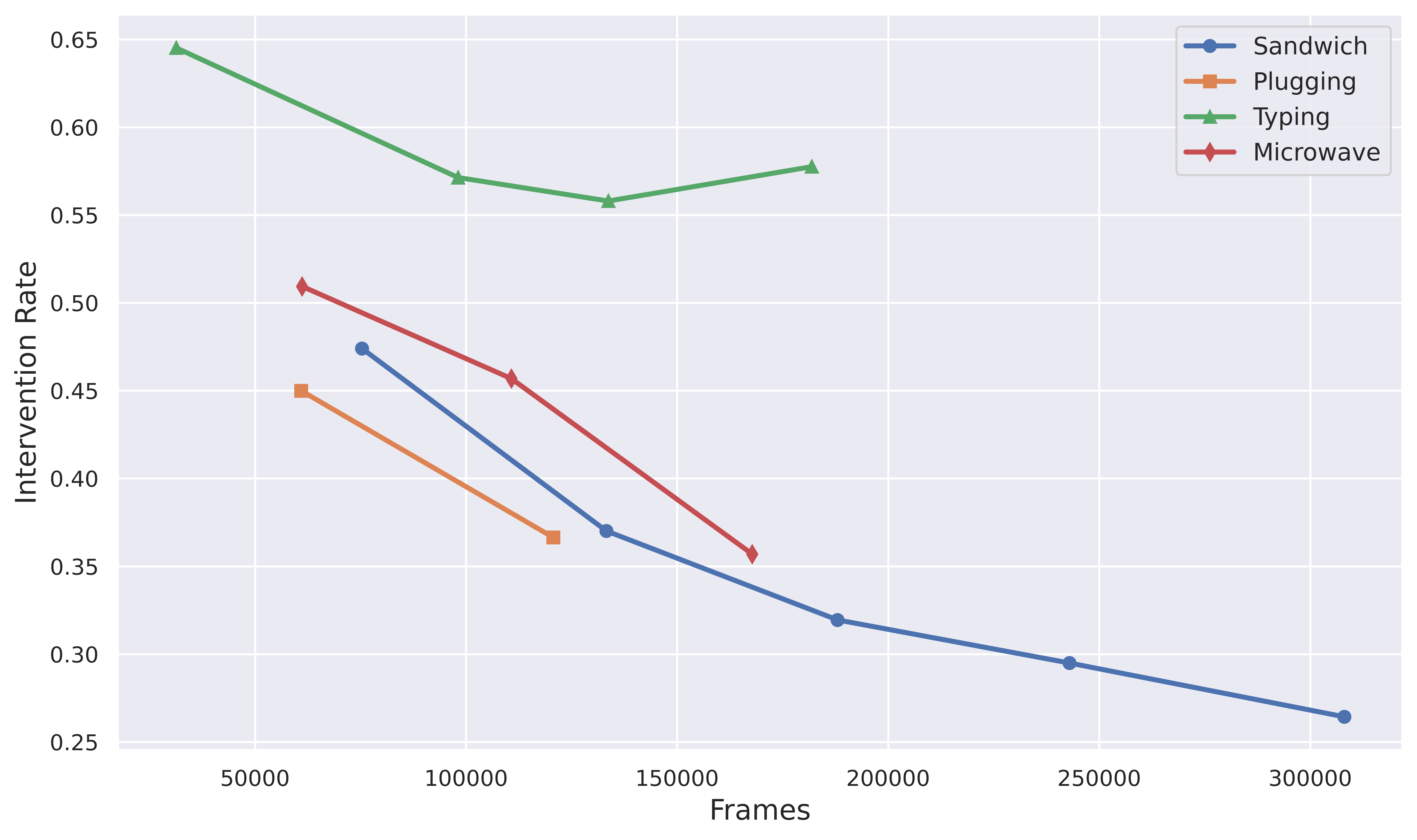}
    \caption{\textbf{Intervention rates} decreased consistently across all tasks as iteration rounds progressed. The typing task, however, exhibited a slight rebound in the final round specifically, as rewind mechanism proportions were intentionally increased only at this terminal stage.}
    \label{fig:takeover}
\end{figure}

\subsection{Q2: Can GCENT Improve Data Efficiency?}

Figure~\ref{fig:efficiency} demonstrates that GCENT achieves comparable or superior performance using only 44.5\% of the frames required by alternative methods. Notably, in the instruction-following task (typing), GCENT attains performance scores exceeding 0.9 with merely 30k frames. We speculate that typing tasks are concentrated and precise operations, so high-quality intervention data for fragile action chunk particularly matters.

As shown in Figure~\ref{fig:takeover}, the human intervention rate, defined as the proportion of frames requiring intervention, decreases significantly over successive GCENT iterations. This reflects continuous policy improvement and a reduced need for human oversight, enabling more efficient parallel supervision of multiple robots.

To assess the scalability of the GCENT framework in single-operator-multi-robot configurations, we conducted a dual-robot experiment with Task Sentinel assistance. As presented in Table~\ref{tab:collection_efficiency}, we evaluated policy models at 40\%, 60\%, and 80\% success rates on the sandwich assembly task, measuring both intervention rates and collection efficiency.

Collected frames represent the combined data collected from both robots. Paused frames measure the waiting time from Task Sentinel's intervention request to operator response. Collection efficiency is calculated by converting the effective frames (collected frames minus paused frames) into a human efficiency ratio, with a maximum of 2.0 for single-operator-dual-robot operation.

Despite a low model success rate of 40\%, the system still achieves a collection efficiency of 1.86, which increases progressively with improved policy performance. This finding establishes that GCENT enhances both data efficiency and quality while substantially reducing human labor costs. These results underscore GCENT's potential for scalable multi-robot supervision and establish a foundation for future extensions to fully concurrent single-operator-N-robot deployments through systematic optimization.

\begin{table*}[t]
\caption{Comparison of data collection efficiency across models with varying success rates under a one-human-two-robot setup in the sandwich task.}
\centering
\resizebox{0.6\textwidth}{!}{%
\begin{tabular}{lccc}
\toprule
\textbf{Metric} & \textbf{40\% Success Rate} & \textbf{60\% Success Rate} & \textbf{80\% Success Rate} \\
\midrule
Intervention Rate (\%) & 47 & 39 & 27 \\
Collected Frames        & 52{,}197 & 61{,}639 & 48{,}523 \\
Paused Frames           & 3{,}746  & 2{,}931  & 1{,}831 \\
Collection Efficiency   & 1.86     & 1.90     & 1.92 \\
\bottomrule
\end{tabular}
}
\label{tab:collection_efficiency}
\end{table*}

\subsection{Q3: How do rewind strategies make impact?}

We observed that the rewind mechanism produces different effects at various stages of GCENT. During initial stages, direct intervention without rewinding enables the model to develop failure recovery capabilities, thereby enhancing overall task performance. In subsequent stages, after achieving higher success rates, activating the rewind mechanism substantially improves first-attempt success rates by eliminating redundant error sequences and refining trajectory quality.

To validate our findings, we conducted controlled experiments on the sandwich assembly task using models with initial success rates of 20\% and 80\%. We compared two intervention strategies: (1) \textit{Direct Intervention} and (2) \textit{Rewind}. As presented in Table~\ref{tab:rewind_results}, the results strongly support our observations: during early stages, direct intervention strategy enables models to learn from corrective demonstrations and enhance robustness through failure recovery mechanisms. In later stages, the rewind strategy facilitates optimal trajectory learning, reduces error attempts, and improves both efficiency and final performance.

\begin{table}[ht]
\caption{Comparison of rewind strategies across training stages in the sandwich assembly task. Direct intervention demonstrates superior performance during early stages, while rewind enhances performance in later stages.}
\centering
\begin{tabular}{lcc}
\toprule
\textbf{Rewind Strategy} & \textbf{Start at 20\%} & \textbf{Start at 80\%} \\
\midrule
Direct Intervention      & \textbf{0.69 $\pm$ 0.07} & 0.84 $\pm$ 0.03 \\
Rewind         & 0.50 $\pm$ 0.06 & \textbf{0.91 $\pm$ 0.01} \\
\bottomrule
\end{tabular}
\label{tab:rewind_results}
\end{table}

\subsection{Fine-tune Details}

To optimize computational efficiency and accelerate convergence, we implemented three methodological refinements to the training pipeline. First, we computed inter-frame joint angle differentials and established a minimal motion threshold ($\pi$/180/30 radians). Frames exhibiting sub-threshold angular displacement were classified as static and subsequently excluded from the training data.

Second, we transformed raw joint angle measurements into end-effector pose representations. Specifically, we derived the differential pose between consecutive frames based on forward kinematics. This relative pose-based representation provided more precise action parameterization for robotic manipulation tasks.

Finally, we implemented dimension-wise min-max normalization on all action components, constraining values to the [-1, 1] interval. This standardization mitigated the adverse effects of heterogeneous scaling across action dimensions, promoting gradient stability during backpropagation and facilitating more efficient optimization dynamics throughout the training process.

%===============================================================================

\section{Conclusion}
\label{sec:conclusion}
Training high success rate and deployable real-world robot policies, particularly Vision-Language-Action (VLA) models, faces a major bottleneck in data collection. Traditional teleoperation is costly, inefficient, and difficult to scale. To address this, we introduce GCENT, a scalable training paradigm for real-world robot policy deployment. GCENT introduces human rewind and refine guidance, of which human operators intervene only upon failure, and a rewind mechanism restores the robot to a valid prior state, to collect corrective demonstrations focused on failure recovery.

Empirical results show that GCENT improves both data efficiency and final task success rates by over 40\% compared to state-of-the-art collection methods. We further propose a Task Sentinel mechanism that allows the model to autonomously detect potential failures and proactively request human intervention, thereby significantly reducing the need for laborious human monitoring. As the policy improves, the frequency of intervention declines, ultimately enabling efficient 1-to-N supervision, where a single human can oversee multiple robots. This is crucial for a scalable data collection system.

In summary, GCENT provides a practical, efficient, and scalable framework for training high-performance, deployable robot policies. Its scalability based on Task Sentinel has the potential to significantly lower the cost of robot learning and accelerate the real-world deployment of intelligent robotic systems.

Future work will develop systems for larger-scale deployments and utilize post-training algorithms to better leverage these negative examples.

% \section{Limitations and Future Work}

% \textbf{Limitations.} While our current framework is largely automated, manual involvement is still required for step annotation and verification. Our multi-robot control system presents operational challenges, requiring complex initialization procedures before each teleoperation session. The rewind mechanism can only handle tasks that can be recovered. Future work will develop systems for larger-scale deployments. Additionally, we have not thoroughly investigated the failed trajectories collected by \projname; future studies will utilize post-training algorithms to better leverage these negative examples.

%===============================================================================
% \clearpage

\bibliographystyle{IEEEtran}
\bibliography{IEEEabrv,reference}

\end{document}